\RequirePackage{fix-cm}
\documentclass[twocolumn]{svjour3}

\usepackage{graphics} 
\usepackage{epsfig} 
\usepackage{mathptmx} 
\usepackage{times} 
\usepackage{amsmath} 
\usepackage{amssymb}  
\usepackage{csvsimple}
\usepackage{array}
\usepackage{float}
\usepackage{verbatim}
\usepackage{subfig}
\usepackage{makecell}
\usepackage{hyperref}

\smartqed  

\usepackage[]{algorithm}
\usepackage[noend]{algpseudocode}
\title{
Robust GNSS Denied Localization for UAV Using Particle Filter and Visual Odometry
}

\author{Rokas Jurevi\v{c}ius          \and
        Virginijus Marcinkevi\v{c}ius \and 
        Justinas \v{S}eibokas
}

\institute{R. Jurevi\v{c}ius \at
              Vilnius University Institute Of Data Science And Digital Technologies\\  
              Akademijos str. 4, Vilnius LT-08663, Lithuania\\
              \email{rokas.jurevicius@mii.vu.lt}
                         \and
           V. Marcinkevi\v{c}ius \at
              Vilnius University Institute Of Data Science And Digital Technologies\\  
              Akademijos str. 4, Vilnius LT-08663, Lithuania\\
              \email{virginijus.marcinkevi\v{c}ius@mii.vu.lt}
              \and
           J. \v{S}eibokas \at
              Vilnius University Faculty of Mathematics and Informatics\\  
              Didlaukio g. 47, LT-08303 Vilnius\\
              \email{justinas.seibokas@mif.stud.vu.lt}
}

\begin{document}

\maketitle
\thispagestyle{empty}
\pagestyle{empty}

\begin{abstract}

Conventional autonomous Unmanned Air Vehicle (abbr. UAV) autopilot systems use Global Navigation Satellite System (abbr. GNSS) signal for navigation. However, autopilot systems fail to navigate due to lost or jammed GNSS signal. To solve this problem, information from other sensors such as optical sensors are used. Monocular Simultaneous Localization and Mapping algorithms have been developed over the last few years and achieved state-of-the-art accuracy. Also, map matching localization approaches are used for UAV localization relatively to imagery from static maps such as Google Maps. Unfortunately, the accuracy and robustness of these algorithms are very dependent on up-to-date maps. The purpose of this research is to improve the accuracy and robustness of map relative Particle Filter based localization using a downward-facing optical camera mounted on an autonomous aircraft. This research shows how image similarity to likelihood conversion function impacts the results of Particle Filter localization algorithm. Two parametric image similarity to likelihood conversion functions (logistic and rectifying) are proposed. A dataset of simulated aerial imagery is used for experiments. The experiment results are shown, that the Particle Filter localization algorithm using the logistic function was able to surpass the accuracy of state-of-the-art ORB-SLAM2 algorithm by 2.6 times. The algorithm is shown to be able to navigate using up-to-date maps more accurately and with an average decrease of precision by 30\% using out-of-date maps.

\end{abstract}


\section{INTRODUCTION}

Autopilot systems fail to navigate due to lost or jammed GNSS signal. This is an important issue in the military field, where any kind of radio communications could be jammed and space missions where GNSS signal is not available. Terrain Contour Matching (abbr. TERCOM) system was developed for cruise missile navigation before the GNSS system was widely available \cite{golden1980terrain}. TERCOM navigation systems are still in use, but they are dependent on the latest height maps, which are not universally available. Digital Scene Mapping and Area Correlation (abbr. DSMAC) was an improved system with the same idea of TERCOM, but instead of heights, it uses aerial imagery obtained using optical camera sensor, but it is also very dependent on up-to-date maps \cite{kendoul2012survey}. DSMAC systems are very sensitive to shadows, so the maps must be made at the same time of the day as the navigation is performed. DSMAC systems were proven to be useful and used for lunar module landing in the Apollo missions. \\
Recent developments of Visual SLAM algorithms have achieved high precision localization in real-time. Visual SLAM algorithm developed in \cite{weiss2011monocular} was shown to achieve centimeter-level precision in an in-doors environment. ORB-SLAM2 and LSD-SLAM achieved accurate results on TUM-RGBD and KITTI datasets \cite{mur2015orb} \cite{engel2014lsd}, however, these datasets provide forward-facing camera images only. SLAM techniques suffer from increasing localization error due to the lack of a global localization. It would be important to evaluate ORB-SLAM2 and LSD-SLAM localization accuracy using downward-facing camera images. 
Visual odometry algorithms are also proposed to solve this problem, although they achieved very high performance and precision, they also suffer from increasing error over long-distance flights.\\
A map matching localization approach is presented in \cite{shan2015google}, where a UAV is localized relatively to imagery from Google maps. During a test flight, algorithms achieved root mean square error of around 6 meters; unfortunately, the authors do not provide absolute mean error, which could be used for an actual comparison against other methods.
This paper analyzes the effects of using different mathematical functions to convert image similarity to particle likelihood value. Image similarity is calculated using the Pearson Correlation Coefficient \cite{pearson1896mathematical} between map and an aerial image obtained from the UAV camera. Pearson Correlation Coefficient provides an arbitrary similarity value that is proportional to image similarity in range of $[-1; 1]$, where one means that images are identical and -1 means that one image is completely anti-correlated. The similarity cannot be used to sample a Particle set, which is the basic idea of the Particle Filtering algorithm. The similarity value must be recalculated to a probability distribution, which can be sampled. The conversion is done in two steps: at first image similarity value using likelihood conversion function (see section \ref{sec:conversion_fns}) is converted to a positive value and secondly using normalization of these values is converted to probability mass (a discrete probability distribution). To calculate probability from image similarity, two parametric conversion functions are proposed, and their impact on accuracy and robustness is measured.

Source code used in these experiments is open-sourced and is available at \footnote{\url{https://github.com/jureviciusr/particle-match}}.
Section \ref{sec:localization_theory} provides an overview of the Particle Filter localization algorithm, proposed conversion functions, and dataset that is used in the experiments. Section \ref{sec:experimental} describes research methodology and provides experimental results. Final section \ref{sec:conclusions} summarizes conclusions of the research.

\section{Optical Terrain Relative Localization}
\label{sec:localization_theory}

The main objective of this research is to calculate map relative location of the UAV, using a downward-facing camera mounted on the aircraft body. Imagery from the camera is matched to an orthophoto map during the flight, in case the GNSS is lost or unavailable. This section provides the background of the Particle Filter localization algorithm, introduces the problem of converting image similarity to probability, and proposes conversion functions to deal with the problem. 

\subsection{Particle Filter Localization}

\algdef{SE}[DOWHILE]{Do}{doWhile}{\algorithmicdo}[1]{\algorithmicwhile\ #1}%

\begin{algorithm*}
\caption{Proposed Particle Filter Localization Algorithm}\label{alg:particle_filter}
\begin{algorithmic}
\State Inputs: Set of particles $ S_{t-1} $ obtained from previous iteration, Latest camera image $ T $, Conversion function parameter $ v $ (if applicable)
\State Outputs: Proposed location $ <x', y', \theta'_{yaw}> $ of the UAV in map coordinates
\Function{FilterParticles}{$S_{t-1}$, $T$, $v$} 
  \State $ H = 0, k = 1, S_t = \emptyset, i = 0 $
  \Do
    \State $ P^{(i)} \sim S_{t-1} $ \Comment{Sample a particle from the particle set}
	\State $ PropagateParticle(P^{(i)}) $ \Comment{Propagate particle using planar motion model equations \ref{eq:mm1}, \ref{eq:mm2}, and \ref{eq:mm3}}
    \State $I = ExtractMapImage(P^{(i)})$ \Comment{Extract map image corresponding to particle location}
    \State $ r_t^{(i)} = CalcSimilarity(T, I) $ \Comment{Calculate image similarity using NCC metric}
    \State $ s_t^{(i)} = F(r_t^{(i)}, v) $ \Comment{Convert image similarity value using logistic conversions function}
    \State $ H = H + s_t^{(i)} $ 
	\State $ S_t = S_t \cup {P^{(i)}} $ \Comment{Insert particle into the particle set}
    \If {$P^{(i)}$ falls into empty bin}
    	\State $ bin = $ non-empty \Comment {Mark bin as non-empty}
        \State $ k = k + 1 $ \Comment {Increase marked bin counter}
    \EndIf
    \State $ i = i + 1 $ \Comment{Increase number of particles}
  \doWhile{$i < \frac{1}{2\epsilon}Z_{k-1,1-\delta}^2$} \Comment{Until K-L bound is reached}
  \State $ n = i $ \Comment{Save the number of particles}
  \State $ x' = 0, y' = 0, \theta'_{yaw} = 0$
  \For{i = 1 .. n}
      \State $ b_t^{(i)} = s_t^{(i)} / H $ \Comment{Calculate particle weight}
	  \State $ x' = x' + (b_t^{(i)} * x_t^{(i)}) $ \Comment{Calculate weighted sum of particle coordinates}
	  \State $ y' = y' + (b_t^{(i)} * y_t^{(i)}) $ \Comment{$x_t^{(i)}, y_t^{(i)}$ are $i$th particle's coordinates in map}
      \State $ \theta'_{yaw} = \theta'_{yaw} + (b_t^{(i)} * \theta_t^{(i)})$ \Comment{Final pose heading angle $\theta_{yaw}$}
  \EndFor
  \Return $<x', y', \theta'_{yaw}>$
\EndFunction
\end{algorithmic}
\end{algorithm*}

Localization is achieved by sampling a set of particles using their probability distribution. Particles are hypothesized UAV locations on the map and are assigned a likelihood value that is proportional to the likeliness of being the true location. Particles are iteratively re-sampled using sampling technique which adapts sampled particle count $n$ depending on Kueller-Leiblach distance (abbr. KLD sampling):
\begin{align}
\label{eq:kld}
n = \frac{1}{2\epsilon}Z^2_{k-1,1-\delta} = \frac{k - 1}{2\epsilon}\Bigg\{1-\frac{2}{9(k-1)} + \sqrt[]{\frac{2}{9(k-1)}}z_{1 - \delta}\Bigg\}^3,
\end{align}
where $z_{1-\delta}$ is the upper $1-\delta$ quantile of the standard normal $N(0, 1)$ distribution and $\epsilon$ is the upper bound of the discrete distribution \cite{fox2002kld}. The technique has shown good results against other sampling techniques on simulated flight data — it provides the same localization accuracy, but dynamic particle count allows to decrease computational costs up to 1.7 times \cite{jurevicius2016comparison}. The particles can be sampled only if the probability distribution is known; the problem is how probability distribution could be calculated from image similarity. Particle Filter uses a finite amount of particles; thus, a probability mass function is going to be used as a discrete probability distribution. 
Algorithm \ref{alg:particle_filter} shows the Particle Filter localization. The initial particle set at time $t$ $S_0$ is generated around the starting point within 300 meters radius and initial likelihood is set to $1$. During the first iteration, all particles have an equal likelihood to be sampled. The bins used in the algorithm is implemented by dividing map coordinates into a 2-D grid of 5 meters. If a particle falls into a bin (a grid square), it is marked as taken. The number of bins is used in the Kueller-Leiblach distance calculation to adapt the number of particles according to their distribution on the map. Image similarity value is calculated using the Pearson Correlation Coefficient value $r$ between the image from the UAV and the map image corresponding to the particle location. Pearson Correlation Coefficient is calculated using these equations:
\begin{align}
\label{eq:corrcoeff}
&r = \frac{\sum_{x=0}^{W}\sum_{y=0}^{H}I'(x, y) \cdot T'(x,y)}{\sqrt{\sum_{x=0}^{W}\sum_{y=0}^{H}I'(x, y)^2} \cdot \sqrt{\sum_{x=0}^{W}\sum_{y=0}^{H}T'(x,y)^2}},\\
&I'(x, y) = I(x, y) - \frac{\sum_{x=0}^{W}\sum_{y=0}^{H}I(x, y)}{W \cdot H},\\
&T'(x, y) = T(x, y) - \frac{\sum_{x=0}^{W}\sum_{y=0}^{H}T(x, y)}{W \cdot H},
\end{align}
where
\begin{itemize}
\setlength\itemsep{0em}
\item $I$ is the gray scale image from the camera,
\item $T$ is the gray scale image from the map according to a particle location,
\item $x, y$ are pixel coordinates,
\item $W, H$ are image dimensions, width and height accordingly.
\end{itemize}
If the images are colored, they are converted to gray scale images. The correlation value $r$ is converted to likelihood using one of the conversion functions $F(x)$ described in section \ref{sec:conversion_fns}. Likelihood can be converted to probability using a normalization: 
\begin{align}
\label{eq:normalization}
b^{(i)} = \frac{F(r^{(i)})}{\Sigma_{y=0}^n F(r^{(y)})},
\end{align}
where 
\begin{itemize}
    \item $r^{(i)}$ is the correlation value of $i$th particle,
    \item $F(r^{(i)})$ is the likelihood of $i$th particle,
    \item $b^{(i)}$ is the probability mass of $i$th particle,
    \item $n$ is the number of particles in the set.
\end{itemize}

The calculated particle probability mass function can be used for sampling in Particle Filter. Sampled particles must be propagated given their motion since the last iteration. Visual Odometry algorithm is used to calculate relative UAV movement over time and is added using a planar motion model. The Visual Odometry algorithm is described in detail in section \ref{sec:odometry}. The particle location update can be calculated from the latest movement data using these equations \cite{thrun2005probabilistic}:

\begin{align}\label{eq:mm1}
x' &= x + \alpha_1 \hat{\delta}_{tran} cos( \theta_{yaw} + \alpha_3, \\
\label{eq:mm2}
y' &= y + \alpha_2 \hat{\delta}_{tran} sin( \theta_{yaw} + \alpha_3\hat{\delta}_{rot}), \\
\label{eq:mm3}
\theta_{yaw}' &= \theta_{yaw} + \alpha_3 \hat{\delta}_{rot},
\end{align}

where
\begin{itemize}
\setlength\itemsep{0em}
\item $ x' $, $ y' $ and $ \theta_{yaw}' $ are the posterior particle location
\item $ \alpha_n $ - measurement noise scale coefficients, selected manually
\item $ \hat{\delta}_{tran} $ - translational (movement speed) measurement with measurement noise $ \epsilon_{tran} $, obtained:\\
\centerline{$ \hat{\delta}_{tran} = \delta_{tran} + sample\_normal(\epsilon_{tran}) $}
\item $ \hat{\delta}_{rot} $ - rotational (heading angle) measurement with measurement noise $ \epsilon_{rot} $, obtained:\\
\centerline{$ \hat{\delta}_{rot} = \delta_{rot} + sample\_normal(\epsilon_{rot}) $}
\item $ sample\_normal $ is Gaussian distribution sampling function:\\
\centerline{$ sample\_normal(\epsilon) = \epsilon \cdot gaussian(0, \frac{1}{3}) $}
\end{itemize}
Final aircraft pose estimation is calculated in a planar 2-dimensional coordinate frame ($X$, $Y$ and flight direction $\theta_{yaw}$) of the map by calculating the weighted sum of the particle locations.

\subsection{Visual Odometry}
\label{sec:odometry}

\begin{figure}[H]
\centering\includegraphics[scale=.7]{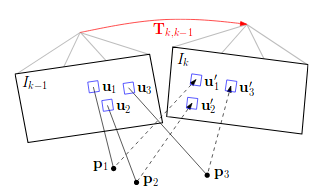}
\caption{Visual odometry calculates motion by aligning 3D world points $p_j$ in image sequence and calculate transformation $T_{k,k-1}$ that causes the features to be reprojected from image patches $u_n$ to $u^\prime_n$ \cite{forster2014svo}}
\label{fig:odometry}
\end{figure}

Visual odometry is the process of calculating aircraft (or robot) motion from the camera image stream. By measuring relative motion between two consecutive frames, it is possible to reduce the search space of the Particle Filter algorithm. The initial particle locations can be propagated using movement measured by a visual odometry algorithm, instead of moving them randomly in the case if no odometry is available. If the propagation step performs an "educated guess" of the posterior particle locations, fewer particles are required; therefore the algorithm can perform faster.\\
Monocular Semi-direct Visual Odometry (abbr. SVO) with a downward-facing camera is used to calculate aircraft motion. SVO algorithm was selected due to more accurate positioning compared to other algorithms and real-time execution on embedded platforms\cite{forster2014svo}. The semi-direct approach used in this algorithm allows very fast execution --- 55 frames per second on embedded computer hardware\cite{forster2014svo}. Figure \ref{fig:odometry} shows the transformation, which is calculated from image sequences. The use of visual odometry reduces the search space of the Particle Filter algorithm.  

\subsection{ORB-SLAM2 Algorithm}

This paper evaluates the Particle Filter localization algorithm against ORB-SLAM2 algorithm, which achieves state-of-the-art results on TUM-RGBD and KITTI datasets \cite{mur2017orb}. The algorithm is available for stereo and mono images, but only mono version will be evaluated since dataset contains only monocular images. The concept of monocular ORB-SLAM2 algorithm is based on ORB feature \cite{rublee2011orb} extraction and matching. Extracted features are matched between two consecutive frames using RANSAC algorithm. The ORB features are tracked in every frame of the image sequence; the camera is localized by matching the features to the local map and minimizing the reprojection error by applying motion only bundle adjustment. The local map is constructed by the algorithm, and the camera is localized relatively to the local map. The output trajectory is scaled according to UAV flight height, and it can be compared against the ground truth.

\subsection{Aerial Imagery Dataset}\label{sec:dataset}

\begin{figure}
\centering\subfloat[Rectangular trajectory over forest map]{\label{main:a}\includegraphics[scale=.15]{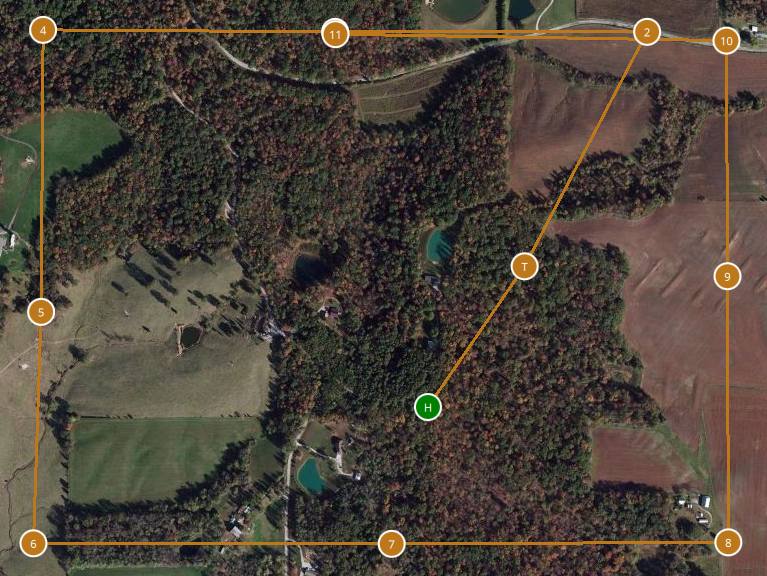}}
\hspace{1mm}
\centering\subfloat[Rectangular trajectory over urban map]{\label{main:b}\includegraphics[scale=.15]{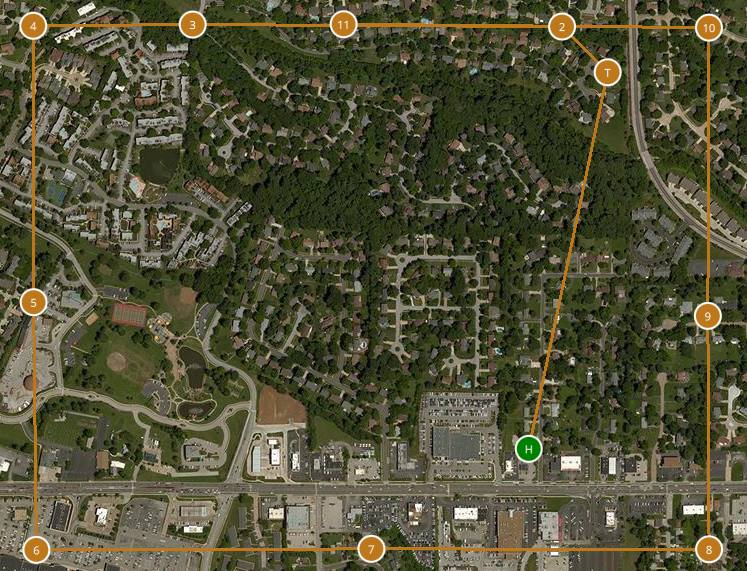}} \\
\centering\subfloat[Circle trajectory over urban map]{\label{main:c}\includegraphics[scale=.15]{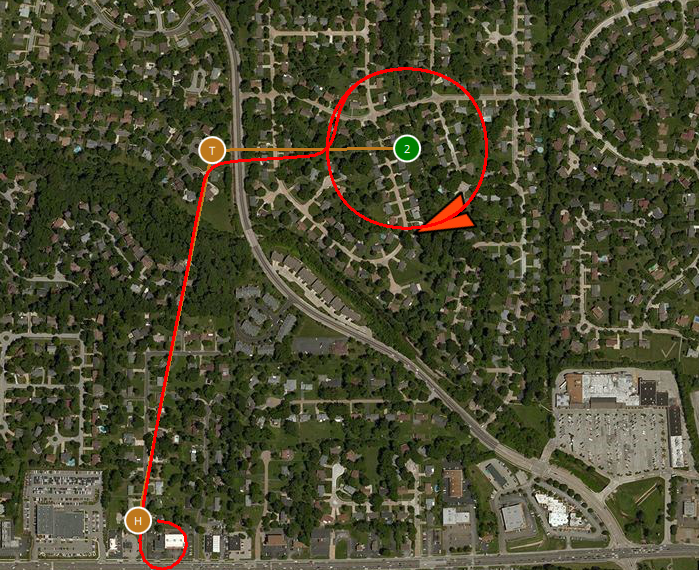}}
\hspace{1mm}
\centering\subfloat[Straight line trajectory over forest map]{\label{main:d}\includegraphics[scale=.147]{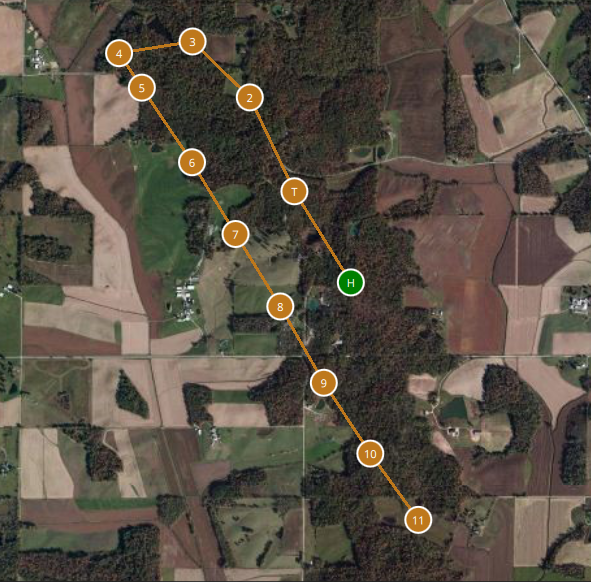}}
\caption{Flight trajectory way points in UAV ground station.}
\label{fig:waypoints}
\end{figure}

This paper presents a new dataset from simulated flights in urban and forested environments using ortho photo maps. Aerial ortho photo imagery was retrieved from USGS website \cite{usdaoffice}. A map is used as ground view in Gazebo simulator environment, running PX4 autopilot software that navigates through the simulated environment and MAVLink message subscribing software, collecting aerial images at 50 frames per second. The PX4 autopilot software is run using software in the loop simulation. Images are recorded alongside with metadata containing aircraft attitude and ground truth location coordinates. Three simple trajectories - straight line, circle, and rectangle were planned at altitudes of 200 and 300 meters over the forest and urban environments, totaling in 12 simulated flights. The dataset is published with open access online \footnote{\url{https://doi.org/10.5281/zenodo.1211729}}. Preview of planned rectangular, circular, and straight-line trajectories are shown in figure \ref{fig:waypoints}. The flight plans are also included in the dataset. QGroundControl software was used to create flight plans. 12 flights are recorded and abbreviated using two characters and a number, e.g., UR-200, where the first character stands for environment (U for urban and F for forest), the second character stands for trajectory (R - rectangle, C - circle, L - straight line) and the number is the altitude of the flight.

\subsection{Image Similarity To Likelihood Conversion Functions}\label{sec:conversion_fns}

This section describes $F(x)$ used for image similarity to likelihood conversion. The easiest approach to calculate likelihood is to convert negative image similarity values to positive. The range of the Pearson Correlation Coefficient is fixed in the range of[-1; 1], the following formula can be used:

\begin{align}
\label{eq:simple_convert}
F(x) = \frac{x + 1}{2}
\end{align}

Equation \ref{eq:simple_convert} provides high likelihoods at poor similarity values, e.g. if $x = -0.1$, then $F(x) = 0.45$. Fig. \ref{conv:linear} shows the visual representation of such conversion. This means that particles with poor image-map similarity survives with high probability. \\
Softmax is a popular function used to convert arbitrary output values to probability densities in neural networks and machine learning \cite{nasrabadi2007pattern}. Softmax conversion function (see figure \ref{conv:softmax}) can be implemented by the following equation :

\begin{align}
\label{eq:softmax}
F(x) = e^x
\end{align}

Due to its wide use amongst machine learning applications, it is going to be used as a baseline in the experimental sections of this paper.\\
The main idea for the conversion function is to achieve robust localization with as little particles as possible. It should assign low probabilities (non-zero) for the negative similarities and boost positive similarities to improve their survivability during sampling. Two functions are proposed to deal with the conversion in the described fashion. The first function uses linear rectification with a single parameter $d$ describing what likelihood value is assigned at $x=0$ using positive parameter value and where the function starts rising on the X-axis using negative parameter values (see figure \ref{conv:prelu}):

\begin{align}
    F(x, d)= 
\begin{cases}
	0, 							& \text{if } d < 0 \text{ and } x \leq |d|\\
    (1 + |d|)(x - |d|) + d^2,	& \text{if } d < 0 \text{ and } x > |d|\\
    d(1 + x),					& \text{if } d\geq 0 \text{ and } x \leq 0\\
    x(1 - d) + d,              	& \text{if } d\geq 0 \text{ and } x > 0\\
\end{cases}
\end{align}

The second function was developed from generalized logistic function \cite{richards1959flexible}:
\begin{align}
l(x) = \frac{A}{(1 + \delta e^{-kx})^\frac{1}{v}}
\end{align}
and by applying fixed coefficients: $A = 1$, $\delta = 1$, $k = 5$, the logistic function becomes:

\begin{align}
L(x, v) = \frac{1}{(1 + e^{-5x})^{\frac{1}{v}}}
\end{align}
 Parameter $v$ is going to be used as a hyper-parameter to control the shape of the curve.
To achieve a curve in range of [0; 1] at input range of [-1; 1], a division is added, so the final conversion equation is:

\begin{align}
F(x, v) = \frac{L(x, v)}{L(1, v)}
\end{align}

See figure \ref{conv:logistic} for the curves with different $v$ values used in the research. This function has a more practical form without different if-cases. It also contains a single hyper-parameter, that is used to control the function curvature.
Particle probability values are calculated by normalizing $F(X)$ using equation \ref{eq:normalization}.
Each of these conversion functions is implemented, and their impact using different hyper-parameters on localization accuracy, speed, and robustness is analyzed.


\newcommand{\myScale}{0.6}
\begin{figure*}
\centering\subfloat[Linear conversion]{\label{conv:linear}\includegraphics[scale=\myScale]{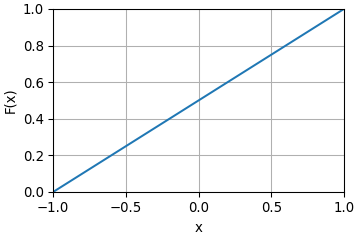}} 
\subfloat[Softmax conversion]{\label{conv:softmax}\includegraphics[scale=\myScale]{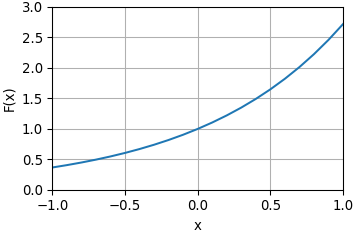}} \\
\centering\subfloat[Rectifying conversion]{\label{conv:prelu}\includegraphics[scale=\myScale]{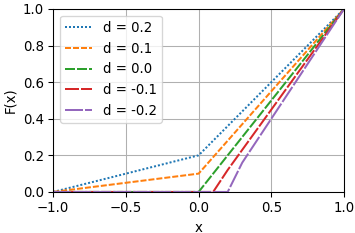}}
\subfloat[Logistic conversion]{\label{conv:logistic}\includegraphics[scale=\myScale]{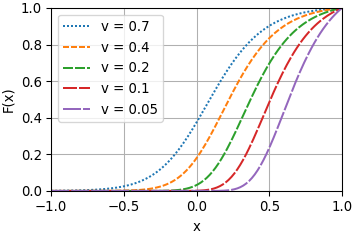}}
\caption{Conversion functions outputs.}
\label{fig:conversion_fns}
\end{figure*}

\section{Experimental results}
\label{sec:experimental}

\subsection{Methodology}

The objective of the experiments is to measures the accuracy and speed of the Particle Filter Localization algorithm using different conversion functions from section \ref{sec:conversion_fns}. Experiments are performed by simulating UAV flights and providing images from the dataset described in section \ref{sec:dataset} as input to the localization algorithm. A single combination of conversion function, a flight scenario, and a map used for matching is run for ten times, and average results are provided to account for randomness in the particle filter algorithm. Maps of different dates are used for matching to test whether the algorithm is able to cope with changes introduced due to the aging of the environment. Four maps are used for forest and urban environment which are created on 2-year intervals. During each experimental flights iteration, a measure of accuracy in meters and duration in particle evaluations are recorded. Rectifying function is used with parameter values: 0.2, 0.1, 0.0, -0.1, -0.2. Logistic function is used with 5 parameter values: 0.7, 0.4, 0.2, 0.1, and 0.05.
Different metrics are used to evaluate accuracy, speed, and robustness:
\begin{itemize}
\item Accuracy is measured by calculating Euclidean distance of dataset ground truth location and algorithm output location in map plane. To select which conversion function gives the most accurate results, the ranking method is used. Average accuracies from each conversion function ranked amongst each of the flight scenarios, given the best - 1 point, 2 points for the second most accurate result and so on, finally, the function with least points will be chosen as the most accurate conversion function.
\item Speed is measured by the number of average evaluated particles during the single experimental flight. Since the number of iterations in each simulated flight is the same, average particle evaluations per iteration gives is proportionate to time of execution, eliminating stochastic changes introduced while measuring execution time and it is independent of system resources.
\item Robustness is evaluated by measuring the average accuracy using maps that were created on a different date. USGS provides imagery of the same regions every two years starting from 2008 (for the regions chosen for this research). This way, we can measure how well the algorithm copes with changes introducing with each map. 
\end{itemize}
The results are also compared to the state-of-the-art ORB-SLAM2 algorithm. It compares localization error of the most accurate Particle Filter and conversion function combination from the ranking against the error of ORB-SLAM2 to see which provides the most accurate results.

\subsection{Accuracy}

\begin{table}
    \caption{Localization accuracy in meters using parametric logistic conversion function\label{tab:prelu_acc}}
    \centering
\begin{tabular} {l r r r r r}%
    \bfseries Scenario 
    & \bfseries v = 0.7
    & \bfseries v = 0.4
    & \bfseries v = 0.2
    & \bfseries v = 0.1
    & \bfseries v = 0.05
    \begin{filecontents*}{accuracy_prelu.csv}
Scenario,d = 0.2,d = 0.1,d = 0.0,d = -0.1,d = -0.2
FL-200,39.27,37.41,55.91,225.41,238.70
FL-300,26.74,26.09,27.66,25.63,160.18
FR-200,62.71,52.73,74.06,62.36,58.20
FR-300,64.34,59.25,31.83,32.54,58.48
FC-200,57.32,68.53,92.39,102.86,49.36
FC-300,95.63,112.06,126.93,156.09,140.20
UL-200,27.17,27.64,26.56,25.46,28.27
UL-300,27.27,25.88,25.95,26.60,27.57
UR-200,54.69,42.43,37.09,66.53,52.23
UR-300,50.00,43.55,41.27,45.07,55.10
UC-200,70.42,56.57,49.09,48.38,65.56
UC-300,73.71,64.74,54.29,63.16,67.49
\end{filecontents*}
\csvreader[head to column names]{accuracy_prelu.csv}{}
    {\\\hline
       \csvcoli   & 
       \csvcolii  &
       \csvcoliii & 
       \csvcoliv  &
       \csvcolv   &
       \csvcolvi}
\end{tabular}
\end{table}

\begin{table}
    \caption{Localization accuracy in meters using parametric rectifying conversion function\label{tab:logistic_acc}}
    \centering
\begin{tabular} {l r r r r r}%
    \bfseries Scenario 
    & \bfseries d = 0.2
    & \bfseries d = 0.1
    & \bfseries d = 0.0
    & \bfseries d = -0.1
    & \bfseries d = -0.2
    \begin{filecontents*}{accuracy_logistic.csv}
Scenario,v = 0.7,v = 0.4,v = 0.2,v = 0.1,v = 0.05 
FL-200,39.97,37.66,50.79,145.09,233.22
FL-300,27.01,25.74,27.14,56.05,184.69
FR-200,59.64,67.15,72.57,72.03,79.35
FR-300,59.50,51.41,40.72,29.18,35.53
FC-200,58.26,73.73,78.47,67.71,26.03
FC-300,92.44,116.12,125.66,142.34,112.88
UL-200,29.81,27.40,26.67,27.03,26.61
UL-300,26.66,26.34,25.01,25.60,25.27
UR-200,58.50,44.85,36.39,42.53,49.48
UR-300,53.54,42.82,39.55,39.92,42.68
UC-200,78.07,72.78,49.43,50.33,48.05
UC-300,76.03,67.38,57.79,56.38,61.53
\end{filecontents*}
\csvreader[head to column names]{accuracy_logistic.csv}{}
    {\\\hline
       \csvcoli   & 
       \csvcolii  &
       \csvcoliii & 
       \csvcoliv  &
       \csvcolv   &
       \csvcolvi}
\end{tabular}
\end{table}

\begin{table}
    \caption{Conversion function ranking according to accuracy\label{tab:ranking}}
    \centering
\begin{tabular} {l r r r}%
    \bfseries Function
    & \bfseries Parameter value
    & \bfseries Score
    & \bfseries Rank
    \begin{filecontents*}{ranking_short.csv}
Function,Parameter value,Score,Place
Rectifying,0.2,75,8
Rectifying,0.1,54,2
Rectifying,0,56,3
Rectifying,-0.1,69,7
Rectifying,-0.2,88,10
Logistic,0.7,82,9
Logistic,0.4,68,6
Logistic,\textbf{0.2},\textbf{52},\textbf{1}
Logistic,0.1,58,4-5
Logistic,0.05,58,4-5
\end{filecontents*}
\csvreader[head to column names]{ranking_short.csv}{}
    {\\\hline
       \csvcoli  
       &\csvcolii  
       &\csvcoliii  
       &\csvcoliv  
       }
\end{tabular}
\end{table}

\begin{table*}
    \caption{Localization accuracy improvements when different functions against Softmax (eq. \ref{eq:softmax}) and linear eq. (\ref{eq:simple_convert}) conversion functions.\label{tab:softmax_comp}}
    \centering
  \begin{tabular} {l r r r r r r}
      \bfseries \thead{}
      & \multicolumn{3}{c}{\textbf{Accuracy improvement}}
      & \multicolumn{3}{c}{\textbf{Speedup}}
      \begin{filecontents*}{acc_comparison.csv}
s,s,s,s,s,s,s
Scenario,Softmax / Linear, Logistic / Linear,Logistic / Softmax,Softmax / Linear,Logistic / Linear,Logistic / Softmax
FL200,7.79\%,15.59\%,7.24\%,1.06,2.41,2.26
FL300,6.47\%,32.40\%,24.35\%,1.07,2.32,2.17
FR200,14.77\%,10.18\%,-4.00\%,1.11,3.23,2.92
FR300,6.37\%,85.07\%,73.99\%,1.14,3.36,2.95
FC200,2.48\%,-40.14\%,-41.59\%,1.00,2.66,2.65
FC300,-2.58\%,-23.61\%,-21.58\%,0.99,2.51,2.53
UL200,12.10\%,65.12\%,47.30\%,1.14,3.76,3.31
UL300,12.37\%,48.47\%,32.12\%,1.12,3.40,3.03
UR200,22.90\%,224.89\%,164.34\%,1.14,3.66,3.21
UR300,6.49\%,98.27\%,86.19\%,1.10,3.37,3.07
UC200,0.81\%,95.25\%,93.67\%,1.01,3.62,3.58
UC300,2.79\%,63.48\%,59.04\%,1.02,3.89,3.83
Average,7.73\%,56.25\%,43.42\%,1.07,3.18,2.96
\end{filecontents*}
\csvreader[head to column names]{acc_comparison.csv}{}
      {\\\hline
         \csvcoli   & 
         \csvcolii  &
         \csvcoliii & 
         \csvcoliv  &
         \csvcolv   &
         \csvcolvi  &
         \csvcolvii
         }
  \end{tabular}
\end{table*}

This section provides localization accuracy results from experiments. Table \ref{tab:prelu_acc} and table \ref{tab:logistic_acc} shows average accuracy results on each flight scenario with logistic and rectifying conversion functions using different parameter values. Table \ref{tab:ranking} shows the experimental results ranked by accuracy for each of the experimental results, by ranking from 1 (best) to 10 (worst), the sum of the ranks is used to determine the most accurate (lowest value) conversion function, and it's parameter value. 
Table \ref{tab:softmax_comp} shows the comparison of selected best conversion function with a parameter value (logistic with v = 0.2) to the baseline Softmax conversion function. The accuracy difference $ \Delta_{accuracy} $ is calculated by dividing the achieved mean localization accuracy $ \overline{\delta}   $ of two functions, e.g. in case of Logistic / Linear: 
\begin{align}
 \Delta_{accuracy} = \Big(\frac{ \overline{\delta}_{linear} - \overline{\delta}_{logistic} }{ \overline{\delta}_{logistic} } \Big) \cdot 100\%.
\end{align}
The comparison results show that the best logistic function is 43\% more accurate and also requires three times fewer iterations on average to achieve the result. 
\subsection{Robustness}
\begin{figure*}
\centering
\subfloat[Accuracies using rectifying conversion function]{\label{orb:rect_robust}\includegraphics[scale=0.4]{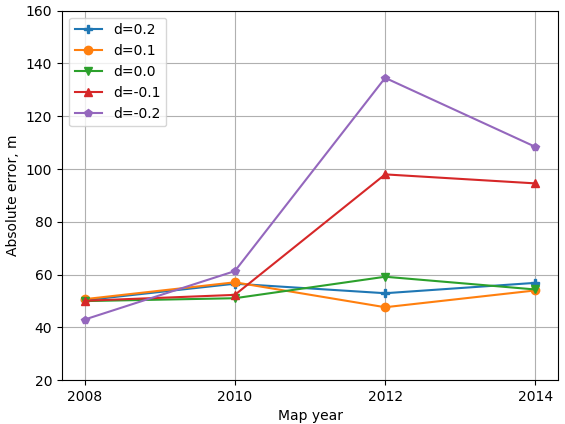}} 
\subfloat[Accuracies using logistic conversion function]{\label{orb:log_robust}\includegraphics[scale=0.4]{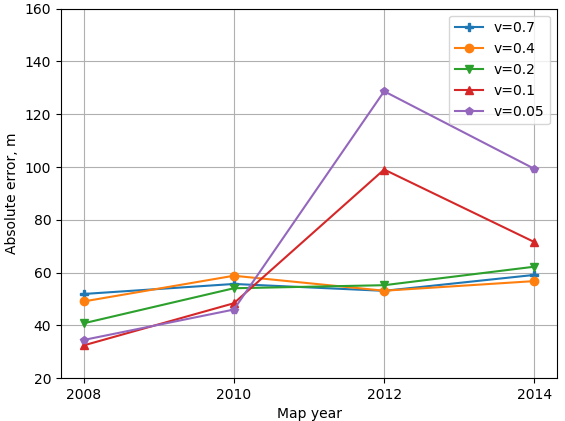}}
\caption{Comparison of localization accuracy with different maps, map year axis depicts the date of map creation.}
\label{fig:conversion_robustness}
\end{figure*}

By comparing average accuracies on different maps, we can evaluate the algorithms ability to localize with changes in the imagery. Figure \ref{fig:conversion_robustness} shows accuracies using different maps for localization. The flight data was collected using 2008 map, so with increasing date of map creation it incorporates more changes. Data shows that using rectifying function with parameter in range $0.0, \dots, 0.2$ and logistic conversion function with parameter in range $0.2, \dots 0.7$ provides similar accuracy with different maps, even with map dated in 2014 which is 6 years apart from the imagery used in the flight. The accuracy also continues to increase with logistic function values $0.1$ and $0.05$ while using very recent maps, but the accuracy drops drastically with maps that were created later in time, so using these values would be provide benefits only if the map is very recent and doesn't have drastic changes.

\subsection{Comparison to ORB-SLAM2}

This section presents the comparison results of the best configuration of the Particle Filter algorithm with state-of-the-art ORB-SLAM2 algorithm. Results of 4 flight scenarios are compared since ORB-SLAM2 was not able to reconstruct other flights and lost position tracking mid-flight; incomplete results are not included. This is most likely due to sudden rotational movements during way-point changes in the trajectory. This might be resolved by increasing framerate of the video, but it was not performed during this research. 
Figures \ref{orb:fl200} to \ref{orb:ul300} shows flight trajectories recovered by 3 algorithms - SVO (which is used internally by Particle Filter), proposed particle filter localization with logistic conversion function and parameter value of 0.2, and ORB-SLAM2. According to results in table \ref{tab:orb_comparison}, ORB-SLAM2 provides similar results to SVO, while being a little bit more precise, meanwhile, Particle Filter localization outperforms both SVO and ORB-SLAM2 with over ~2.6 times higher precision in average.

\newcommand{\myTrajScale}{0.4}
\begin{figure*}[h]
\centering
\subfloat[Forest map, 200 meters altitude]{\label{orb:fl200}\includegraphics[scale=\myTrajScale]{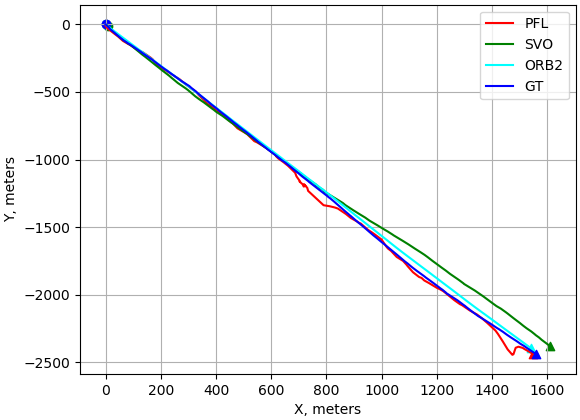}} 
\subfloat[Forest map, 300 meters altitude]{\label{orb:fl300}\includegraphics[scale=\myTrajScale]{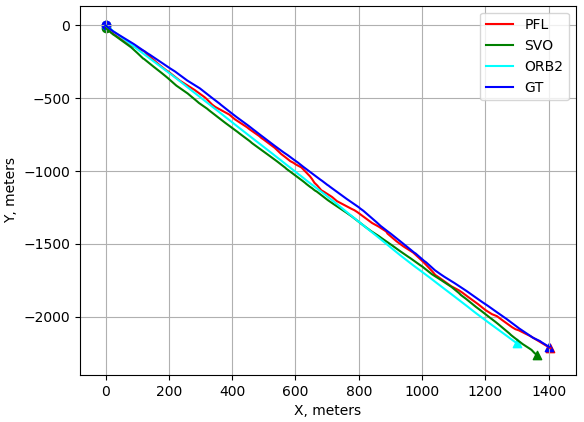}} \\
\subfloat[Urban map, 200 meters altitude]{\label{orb:ul200}\includegraphics[scale=\myTrajScale]{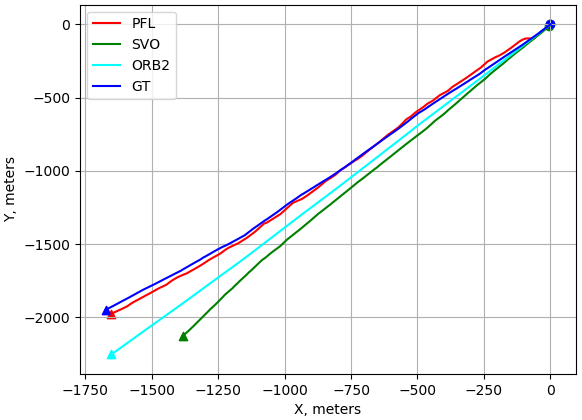}}
\subfloat[Urban map, 200 meters altitude]{\label{orb:ul300}\includegraphics[scale=\myTrajScale]{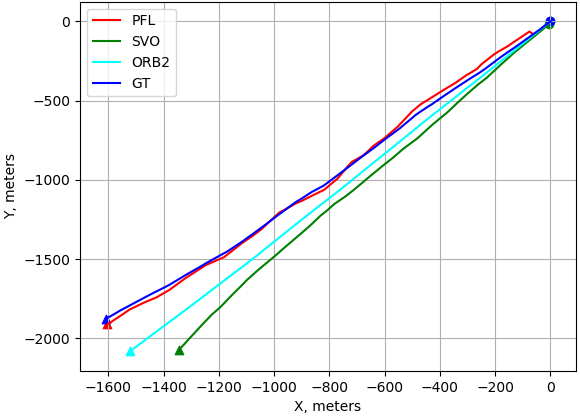}}
\caption{Linear flight trajectories recovered using SVO, ORB-SLAM2, and proposed Particle Filter algorithms. Circle symbol marks start location, triangle symbol marks end location of the flight.}
\label{fig:trajectories}
\end{figure*}

\begin{table*}
    \caption{Localization accuracy comparison of ORB-SLAM2\label{tab:orb_comparison}}
    \centering
\begin{tabular} {l r r r r r r}%
    \bfseries \thead{} 
    & \multicolumn{3}{c}{Accuracy, m}
    & \multicolumn{3}{c}{Relative accuracy}
    
    \begin{filecontents*}{orb_comparison.csv}
Scenario,SVO,PFL,ORB2,PFL/SVO,ORB2/SVO,PFL/ORB2
Scenario,SVO,PFL,ORB2,PFL/SVO,ORB2/SVO,PFL/ORB2
FL-200,48.36,27.60,49.56,75.26\%,-2.43\%,79.61\%
FL-300,45.00,17.08,58.55,163.53\%,-23.14\%,242.88\%
UL-200,130.95,21.93,118.48,497.03\%,10.53\%,440.17\%
UL-300,129.88,21.06,86.05,516.84\%,50.94\%,308.67\%
Average,,,,313.16\%,8.97\%,267.83\%
\end{filecontents*}
\csvreader[head to column names]{orb_comparison.csv}{}
    {\\\hline
       \csvcoli   
       & \csvcolii 
       & \csvcoliii
       & \csvcoliv
       & \csvcolv 
       & \csvcolvi
       & \csvcolvii}
\end{tabular}
\end{table*}

\section{Practical considerations}

This paper proposes a new Particle Filter localization algorithm based on KLD sampling and Pearson correlation coefficient. Few image similarity to likelihood conversion functions are compared, and among those, the logistic function allows to achieve the most accurate results. In 10 flight scenarios out of 12 performed, the logistic function performs more accurately than trivial linear conversion. Additionally, since the logistic function decreases the survival chances of low likelihood particles, the adaptive KLD sampling reduces particle count efficiently, and the algorithm evaluates fewer particles, thus speeding up the execution around three times compared to linear conversion. The logistic function is not the best function to be used in all cases. If the maps are known to be quite different from the expected camera imagery, it is recommended to use either rectifying function with parameter value in range 0.0,...,0.2 or logistic conversion function with parameter value in range 0.2,...,0.7. These configurations are not as accurate, but allows more particles with lower likelihoods to be sampled, increasing robustness of the algorithm to inaccuracies in maps. Logistic function with parameter value 0.2 is recommended for the general case, but parameter value can be decreased to 0.1 or 0.05 to increase accuracy if the maps are known to be up-to-date.

\section{Conclusions}
\label{sec:conclusions}

An application of Particle Filter localization was implemented with different image similarity to likelihood conversion functions. Accuracy, speed, and robustness of localization were measured using different conversion functions and different hyper-parameter values. Over a thousand experimental flights were conducted on the Aerial Imagery dataset to evaluate proposed conversion functions. From all the data collected, we can draw these conclusions:
\begin{itemize}
\item Localization performed using Softmax function (eq. \ref{eq:softmax}) compared with linear conversion function (eq. \ref{eq:simple_convert}) shows that localization using Softmax function improved accuracy error by 7\%.
\item Localization using the proposed logistic function with a parameter value of 0.2 was shown to provide the best results by ranking localization accuracy from all the conversion functions results.
\item Particle Filter localization achieved 43\% higher accuracy using three times fewer computations than the baseline Softmax probability conversion function.
\item Localization using rectifying conversion function with parameter values 0.2, 0.1, and 0.0 and logistic conversion function with parameter values 0.7, 0.4, and 0.2 has shown similar localization accuracy using maps of different dates, showing an ability to be robust to changes in the imagery. Other functions showed higher accuracy with similar maps and inaccurate results using other maps for image matching.

\end{itemize}
The best configuration of the Particle Filter algorithm was compared against state-of-the-art ORB-SLAM2 algorithm. The results show that ORB-SLAM2 provides 9\% higher accuracy than the odometry algorithm SVO. Meanwhile, Particle Filter using the proposed logistic function was able to achieve ~2.6 times better accuracy than both ORB-SLAM2 and SVO. However, the results were only provided for the straight line flight trajectory, since ORB-SLAM2 was unable to complete whole trajectories on the other flight scenarios, while Particle Filter localization completed all flights.

\bibliographystyle{IEEEtran}
\bibliography{IEEEabrv,mybib.bib}

\end{document}